\documentclass{article}

\PassOptionsToPackage{numbers,square}{natbib}

\usepackage[final]{nips_2017}

\usepackage[utf8]{inputenc} 
\usepackage[T1]{fontenc}    
\usepackage{hyperref}       
\usepackage{url}            
\usepackage{booktabs}       
\usepackage{amsfonts}       
\usepackage{amsmath}        
\usepackage{nicefrac}       
\usepackage{microtype}      
\usepackage{enumerate}
\usepackage{subfiles}
\usepackage{xcolor}
\usepackage{multirow}
\usepackage{graphicx}
\usepackage{subfiles}

\newcommand\blfootnote[1]{%
  \begingroup
  \renewcommand\thefootnote{}\footnote{#1}%
  \addtocounter{footnote}{-1}%
  \endgroup
}

\newcommand\cs[2]{ConvStep_{#1}(#2)}
\newcommand\concat[3]{\left[#1 \parallel_#3 #2\right]}

\title{One Model To Learn Them All}

\author{
  {\L}ukasz Kaiser\\
  Google Brain\\
  \texttt{lukaszkaiser@google.com}\\
  \And
  Aidan N. Gomez\thanks{Work performed while at Google Brain.}\\
  University of Toronto\\
  \texttt{aidan@cs.toronto.edu}\\
  \And
  Noam Shazeer\\
  Google Brain\\
  \texttt{noam@google.com}\\
  \And
  Ashish Vaswani\\
  Google Brain\\
  \texttt{avaswani@google.com}\\
  \And
  Niki Parmar\\
  Google Research\\
  \texttt{nikip@google.com}\\  
  \And
  Llion Jones\\
  Google Research\\
  \texttt{llion@google.com}\\   
  \And
  Jakob Uszkoreit\\
  Google Research\\
  \texttt{usz@google.com}\\   
}

\begin{document}

\maketitle

\begin{abstract}
Deep learning yields great results across many fields,
from speech recognition, image classification, to translation.
But for each problem, getting a deep model to work well involves
research into the architecture and a long period of tuning.
We present a single model that yields good results on a number
of problems spanning multiple domains. In particular, this single model
is trained concurrently on ImageNet, multiple translation tasks,
image captioning (COCO dataset), a speech recognition corpus,
and an English parsing task. 
Our model architecture incorporates building blocks from multiple
domains. It contains convolutional layers, an attention mechanism,
and sparsely-gated layers.
Each of these computational blocks is crucial for a subset of
the tasks we train on. Interestingly, even if a block is not
crucial for a task, we observe that adding it never hurts performance
and in most cases improves it on all tasks.
We also show that tasks with less data benefit largely from joint
training with other tasks, while performance on large tasks degrades
only slightly if at all.
\blfootnote{Code available at \url{https://github.com/tensorflow/tensor2tensor}}
\end{abstract}

\section{Introduction}

Recent successes of deep neural networks have spanned many
domains, from computer vision \cite{img12} to
speech recognition \cite{dahl12} and many other tasks.
Convolutional networks excel at tasks related to vision,
while recurrent neural networks have proven successful at
natural language processing tasks, e.g., at machine translation
\cite{sutskever14,bahdanau2014neural,cho2014learning}.
But in each case, the network was designed and tuned specifically
for the problem at hand. This limits the impact of deep learning, as this
effort needs to be repeated for each new task. It is also very different
from the general nature of the human brain, which is able to learn many
different tasks and benefit from transfer learning.
The natural question arises:
\begin{center}
\emph{Can we create a unified deep learning model to solve tasks across multiple domains?}
\end{center}

The question about multi-task models has been studied in many papers
in the deep learning literature. Natural language processing models have been
shown to benefit from a multi-task approach a long time ago \cite{unifiednlp},
and recently machine translation models have even been shown to exhibit zero-shot
learning when trained on multiple langauges \cite{gnmtmulti}.
Speech recognition has also been shown to benefit from multi-task
training \cite{speechmulti}, as have some vision problems, such
as facial landmark detection \cite{facemulti}.
But all these models are trained on other tasks \emph{from the same domain}:
translation tasks are trained with other translation tasks, vision tasks
with other vision tasks, speech tasks with other speech tasks.
Multi-modal learning has been shown to improve learned representations
in the unsupervised setting \cite{multimodaldeepl} and when used as
a-priori known unrelated tasks \cite{unrelatedtasks}.
But no competitive multi-task multi-modal model has been proposed,
so the above question remains unanswered.

In this work, we take a step toward positively answering the above
question by introducing the \emph{MultiModel} architecture,
a single deep-learning model that can simultaneously learn multiple
tasks from various domains. Concretely, we train the MultiModel
simultaneously on the following 8 corpora:
\begin{enumerate}[(1)]
\item WSJ speech corpus \citep{wsj1audio}
\item ImageNet dataset \citep{imagenet}
\item COCO image captioning dataset \citep{COCO}
\item WSJ parsing dataset \citep{marcus1999wsjtreebank}
\item WMT English-German translation corpus
\item The reverse of the above: German-English translation.
\item WMT English-French translation corpus
\item The reverse of the above: German-French translation.
\end{enumerate}

\begin{figure}
  \centering
  \includegraphics[scale=0.55]{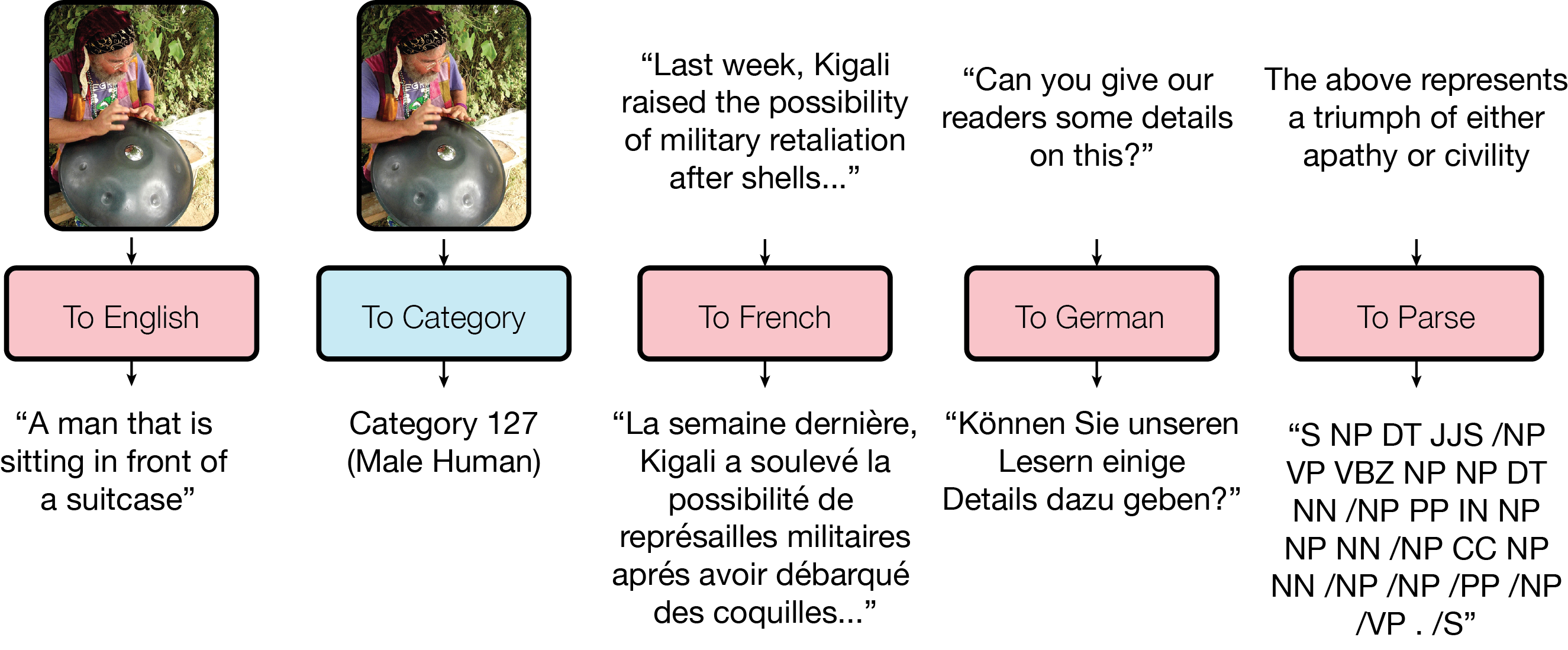}
  \caption{Examples decoded from a single MultiModel trained jointly on 8 tasks. Red depicts a language modality while blue depicts a categorical modality.}
  \label{examplesfig}
\end{figure}

The model learns all of the above tasks and achieves good performance:
not state-of-the-art at present, but above many task-specific
models studied in recent past (see the Section~\ref{sec-experiments} for details).
Figure~\ref{examplesfig} illustrates some decodes taken directly from the model:
it is clear that it can caption images, categorize them,
translate to French and German and construct parse trees.
While the MultiModel is only a first step and will be tuned and improved in the future,
two key insights are crucial to making it work at all and are the main contributions of this work.

\textbf{Small modality-specific sub-networks convert into a unified representation and back from it.} \newline
To allow training on input data of widely different sizes and dimensions,
such as images, sound waves and text, we need sub-networks to convert inputs
into a joint representation space. We call these sub-networks \emph{modality nets}
as they are specific to each modality (images, speech, text) and define transformations
between these external domains and a unified representation.
We design modality nets to be computationally minimal, promoting heavy feature extraction
and ensuring that the majority of computation is performed within the domain-agnostic body
of the model. Since our model is auto-regressive, modality nets need to both convert the inputs
into the unified representation and later convert from this representation into the output space.
Two design decisions were important:
\begin{itemize}
\item \emph{The unified representation is variable-size.}
  While a fixed-size representation is tempting and easier to implement,
  it creates a bottleneck and limits the performance of the model.
\item \emph{Different tasks from the same domain share modality nets.}
  We avoid creating a sub-network for every task, and prefer only to create one for every
  input modality. For example, all translation tasks share the same modality-net (and vocabulary),
  no matter for which language pair. This encourages generalization across tasks
  and allows to add new tasks on the fly.
\end{itemize}

\textbf{Computational blocks of different kinds are crucial for good results on various problems.} \newline
The body of the MultiModel incorporates building blocks from mutiple domains.
We use depthwise-separable convolutions, an attention mechanism, and sparsely-gated mixture-of-experts layers.
These blocks were introduced in papers that belonged to different domains and were not studied before
on tasks from other domains. For example, separable convolutions were introduced in the Xception
architecture \cite{xception2016} and were not applied to text or speech processing before.
On the other hand, the sparsely-gated mixture-of-experts \cite{moe} had been introduced
for language processing tasks and has not been studied on image problems.
We find that each of these mechanisms is indeed crucial for the domain it was introduced,
e.g., attention is far more important for language-related tasks than for image-related ones.
But, interestingly, adding these computational blocks never hurts performance, even on tasks
they were not designed for. In fact we find that both attention and mixture-of-experts layers
slightly improve performance of MultiModel on ImageNet, the task that needs them the least.

\section{MultiModel Architecture}

\begin{figure}
  \centering
  \hspace{0.2cm}
  \includegraphics[scale=0.40]{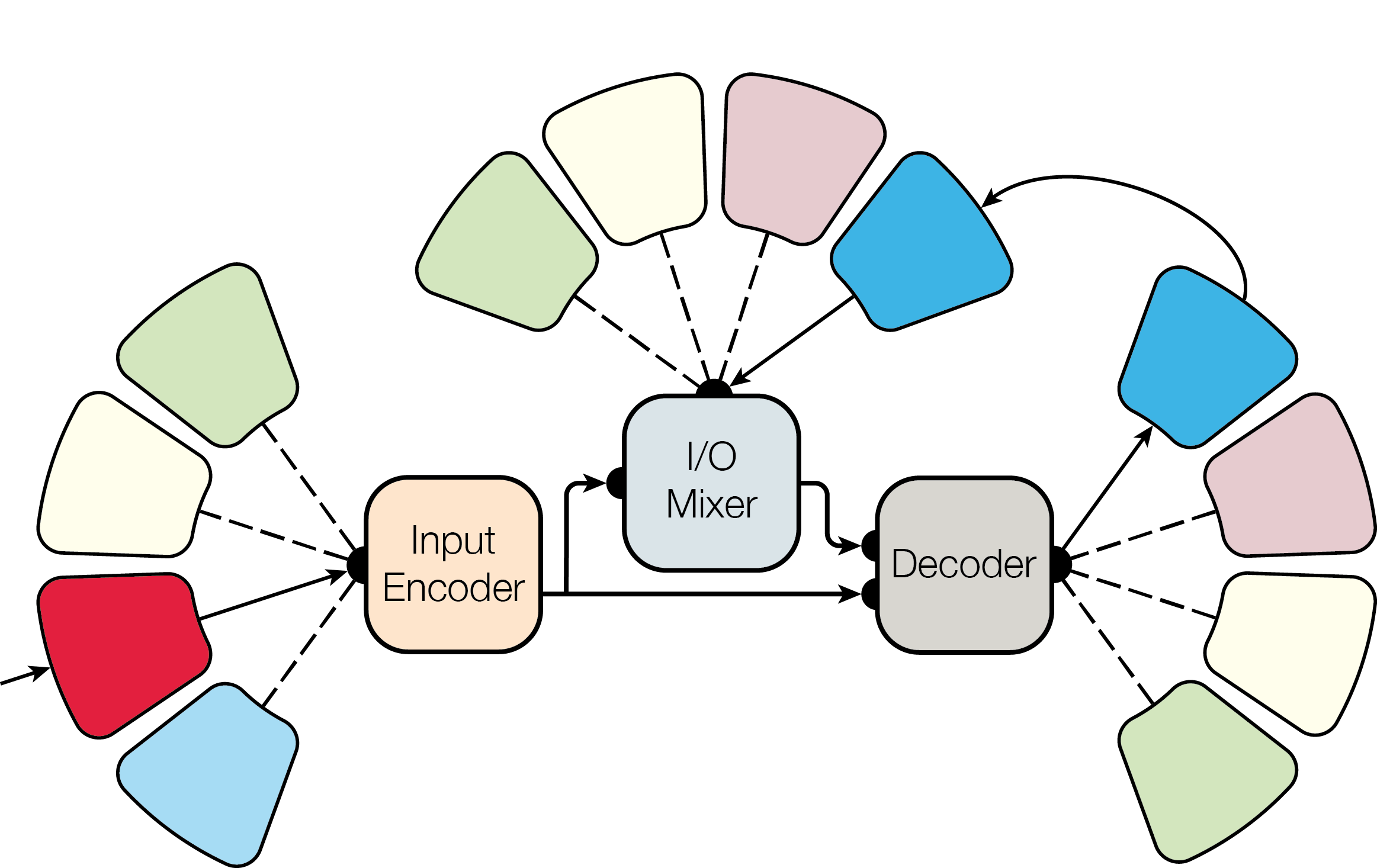}
  \caption{The MultiModel, with modality-nets, an encoder, and an autoregressive decoder.}
  \label{modalityfig}
\end{figure}

The MultiModel consists of a few small modality-nets, an encoder, I/O mixer, and
an autoregressive decoder, as depicted in Figure~\ref{modalityfig}.
As already said above, the encoder and decoder are constructed
using 3 key computational blocks to get good performance across different problems:
\begin{enumerate}[(1)]
    \item Convolutions allow the model to detect local patterns and generalize across space.
    \item Attention layers allow to focus on specific elements to improve performance of the model.
    \item Sparsely-gated mixture-of-experts gives the model capacity without excessive computation cost.
\end{enumerate}
We start by describing the architecture of each of these 3 blocks
and then introduce the encoder, decoder and the architecture of
our modality-nets.

\subsection{Convolutional Blocks}

To perform local computation, we use blocks of convolutions with ReLU non-linearities and normalization.
A block of convolutions gets as input a tensor of shape [\text{batch size}, \text{sequence length}, \text{feature channels}]
and returns a tensor of the same shape, processed as follows.

For convolution operations, we use depthwise separable convolutions, introduced in \cite{xception2016},
in a way similar to \cite{sepconvnmt}.
Depthwise separable convolutions are a parameter- and computationally-efficient variant of the traditional convolution.
They are defined by a convolution on each feature channel separately,
followed by a pointwise convolution to project to the desired feature depth.
We refer the reader to \cite{xception2016} for a complete definition;
here we will denote a depthwise separable convolution with weights $W^{h \times w}$
corresponding to $f$ kernels of size $h \times w$ applied
to an input tensor $x$ with stride $s$ and dilated by a factor $d$ (see \cite{dilatedconv2015})
as $SepConv_{d,s,f}(W, x)$. Note that subscripts for stride, dilation and output size are omitted
when dilation $d$ or stride $s$ are equal to 1,
or output size $f$ is equal to the input's feature depth.

We use convolutions in blocks that consist of three components: a \(ReLU\) activation of the inputs,
followed by a \(SepConv\), followed by layer normalization. 
Layer normalization \cite{layernorm2016} acts over the \(h\) hidden units of the layer below, computing layer-wise statistics for each batch example and normalizing accordingly. These normalized units are then scaled and shifted by scalar learned parameters \(G\) and \(B\) respectively, producing the final units to be activated by a non-linearity.
%
%
The complete convolution step is therefore defined as:
\[ \cs{d,s,f}{W, x} = LN(SepConv_{d,s,f}(W, ReLU(x))). \]

The convolutional steps are composed into blocks by stacking them and adding residual connections
as depicted in Figure~\ref{encdecfig}. We use stacks of four convolutional blocks with two
skip-connections between the stack input and the outputs of the second and fourth convolutional steps,
and with the first two having $3 \times 1$ kernels and the next two having $15 \times 1$ kernels,
with the final one dilated by $8$ to provide a wide receptive field. We also add $40\%$ dropout
at the end of each block, so the complete block is defined as follows:
\begin{align*}
    hidden1(x) &= \cs{}{W^{3 \times 1}_{h1}, x}\\
    hidden2(x) &= x + \cs{}{W^{3 \times 1}_{h2}, hidden1(x)}\\
    hidden3(x) &= \cs{}{W^{15 \times 1}_{h3}, hidden2(x)}\\
    hidden4(x) &= x + \cs{d=8}{W^{15 \times 1}_{h4}, hidden3(x)}\\
    ConvBlock(x) &= \begin{cases}
                Dropout(hidden4(x), 0.4) & \text{during training}\\
                hidden4(x) & \text{otherwise}
              \end{cases}
\end{align*}

\subsection{Attention Blocks}

For attention, we use a multi-head dot-product attention mechanism inspired by \cite{bahdanau2014neural}
and similar to \cite{attnall}, as depicted in Figure~\ref{encdecfig}. The inputs to the attention layer are two tensors: a \(source\) tensor
and a \(target\) tensor both with the shape \([\text{batch size}, \text{sequence length}, \text{feature channels}]\)
The \(target\) tensor is additively composed with a timing signal and mixed using two convolutional blocks. This mixed tensor is then self-attended using a multi-head dot-product attention, which is a dot-product attention with inputs split into \(g=8\) separate tensors representing each attention head, as shown in Figure~\ref{encdecfig}.
The timing signals are the main difference between this attention mechanism and the ones used previously.
They allow this content-based attention to focus based on their position. They are constructed
by concatenating sine and cosine curves:
\begin{align*}
    \Delta(2d) &= 1e4^{-\frac{2d}{depth}} \\
    timing{(t, [2d,2d+1])}   &= \concat{\sin(t\Delta(2d))}{\cos(t\Delta(2d))}{2}
\end{align*}
where \([a||_db]\) represent concatenation of \(a\) and \(b\) along the \(d^{\text{th}}\) dimension.
The \(source\) tensor is finally passed through two different pointwise convolutions to generate the memory keys \(K\) and values \(V\) and the query keys, memory keys and memory values are used to apply the attention mechanism between the self-attended \(target\) and the \(source\) (see Figure~\ref{encdecfig}).

%

\subsection{Mixture-of-Experts Blocks}

We use sparsely-gated mixture-of-experts layers of the same kind as introduced in \cite{moe}:
A mixture-of-experts layer consists of a number of simple feed-forward neural networks (experts)
and a trainable gating network which selects a sparse combination of the experts to process each input.
We refer the reader to \cite{moe} for details as we use exactly the architecture described there.
In particular, during training we select $k=4$ experts out of the whole expert pool and add
the additional load-balancing cost as in \cite{moe}. In each of the two mixture-of-experts
layers in our model, we use a pool of $240$ experts when training on 8 problems jointly,
and $60$ experts when training on each problem separately.

\subsection{Encoder and Mixer and Decoder}

The body of the MultiModel consists of 3 parts:
the encoder that only processes the inputs,
the mixer that mixes the encoded inputs with
previous outputs (autoregressive part),
and a decoder that processes the inputs and
the mixture to generate new outputs.

\begin{figure}[ht]
  \centering
  \includegraphics[scale=0.28]{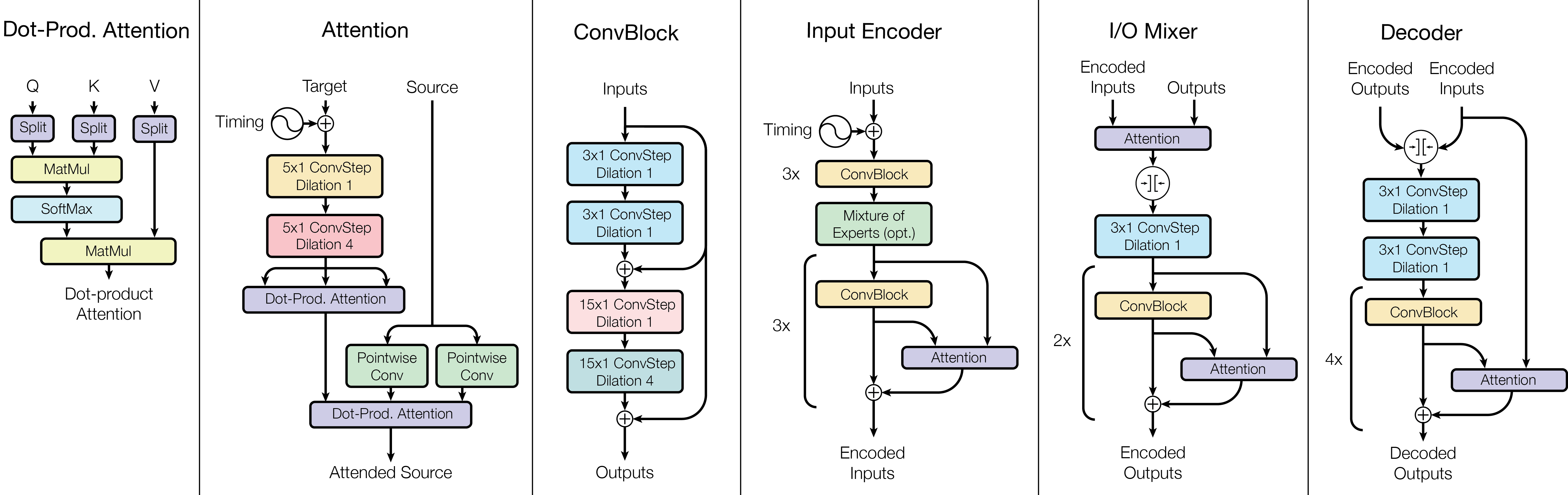}
  \caption{Architecture of the MultiModel; see text for details.}
  \label{encdecfig}
\end{figure}

The encoder, mixer and decoder are structured similarly to previous
fully convolutional sequence to sequence models such as
ByteNet~\cite{bytenet2016} or WaveNet~\cite{wavenet2016},
but differ in the computational blocks that are used.
We depict their architecture in Figure~\ref{encdecfig}.
As can be seen there, the encoder consists of 6 repeated
convolutional blocks (described before) with a mixture-of-experts
layer in the middle.
The mixer consists of an attention block and 2 convolutional
blocks. The decoder consists of 4 blocks of convolutions and
attention, with a mixture-of-experts layer in the middle.
Crucially, the convolutions in the mixer and decoder
are padded on the left, so they can never access any information in
the future. This allows the model to be autoregressive,
and this convolutional autoregressive generation scheme offers
large receptive fields over the inputs and past outputs,
which are capable of establishing long term dependencies.

To allow the decoder to produce outputs for different tasks
even with the same modality, we always start decoding with
a command-token, such as \emph{To-English} or \emph{To-Parse-Tree}.
We learn an embedding vector corresponding to each of the tokens
during training.

%

\subsection{Modality Nets}

We have 4 modality nets, for language (text data), images, audio, and categorical data.

\subsubsection{Language modality net}

Our language-based data is all tokenized using the same vocabulary with 8k subword-units,
following the method from \cite{SennrichHB15}.
The language input modality takes a sequence of tokens ending in a termination token. This sequence of tokens is mapped to the correct dimensionality for the body using a learned embedding.
On the output side, the language modality takes the decoded output of the body and performs a learned linear mapping,
followed by a \(Softmax\), resulting in a probability distribution over the token vocabulary.
\begin{align*}
    LanguageModality_{\text{in}}(x, W_E) &= W_E \cdot x\\
    LanguageModality_{\text{out}}(x, W_S) &= Softmax(W_S \cdot x)
\end{align*}

\subsubsection{Image modality net}\label{imagemodality}

The image input modality is analogous to the Xception entry flow \cite{xception2016}.
The input image's feature depth is gradually deepened using residual convolution blocks
which we call \(ConvRes\) and define as follows:
\begin{align*}
    c1(x,F) &= \cs{f=F}{W^{3 \times 3}, x}\\
    c2(x,F) &= \cs{f=F}{W^{3 \times 3}, c1(x,F)}\\
    p1(x,F) &= MaxPool_2([3 \times 3], c2(x,F))\\
    ConvRes(x, F) &= p1(x, F) + \cs{s=2}{W^{1 \times 1}, x},
\end{align*}
where \(MaxPool_s([h \times w], x)\) is a max-pooling layer over \(x\) with stride \(s\) and window shape \([h \times w]\).
The ImageModality input flow with network depth $d$ (we use $d = 1024$) is defined as:
\begin{align*}
    h1(x) &= \cs{s=2,f=32}{W^{3 \times 3}, x}\\
    h2(x) &= \cs{f=64}{W^{3 \times 3}, h1(x)}\\
    r1(x) &= ConvRes(h2(x), 128)\\
    r2(x) &= ConvRes(r1(x), 256)\\
    ImageModality_{\text{in}}(x) &= ConvRes(r2(x), d)
\end{align*}

\subsubsection{Categorical modality net}

The categorical output modality is analogous to the Xception exit flow \cite{xception2016}.
If the network inputs are two-dimensional data such as image or spectral audio data,
then the one-dimensional output from the model body is first reshaped into two-dimensions again,
followed by progressive down-sampling:
\begin{align*}
    skip(x) &= \cs{s=2}{W^{3 \times 3}_{skip}, x}\\
    h1(x) &= \cs{}{W^{3 \times 3}_{h1}, x}\\
    h2(x) &= \cs{}{W^{3 \times 3}_{h2}, h1(x)}\\
    h3(x) &= skip(x) + MaxPool_2([3 \times 3], h2(x))\\
    h4(x) &= \cs{f=1536}{W^{3 \times 3}_{h4}, h3(x)}\\
    h5(x) &= \cs{f=2048}{W^{3 \times 3}, h4(x)}\\
    h6(x) &= GlobalAvgPool(ReLU(h5(x)))\\
    CategoricalModality_{\text{out}}(x) &= PointwiseConv(W^{classes}, h6(x))
\end{align*}

\(GlobalAvgPool\) denotes a mean taken across all spatial and temporal dimensions.

\subsubsection{Audio modality net}

We accept audio input in the form of a 1-dimensional waveform over time or as a 2-dimensional spectrogram. Both the waveform and spectral input modalities use a stack of 8 \(ConvRes\) blocks from the \(ImageInputModality\) (Section \ref{imagemodality}). The \(i^{\text{th}}\) block has the form: \(l_i = ConvRes(l_{i-1}, 2^i)\). The spectral modality does not perform any striding along the frequency bin dimension, preserving full resolution in the spectral domain.

\subsection{Related Models}

The MultiModel architecture draws from eariler encoder-decoder
architectures applied to neural machine translation.
Earlier sequence-to-sequence models for translation \cite{sutskever14,bahdanau2014neural,cho2014learning}
used recurrent neural networks (RNNs) with long short-term
memory cells \cite{hochreiter1997}).
Convolutional architectures yielded good results on word-level
neural machine translation starting from \cite{KalchbrennerB13}
and later in \cite{MengLWLJL15}. These early models used a standard
RNN on top of the convolution to generate the output and had
a bottleneck there that hurt performance, especially on longer
sentences, similarly to  the limitations of RNN
sequence-to-sequence models without attention \cite{sutskever14,cho2014learning}.
Fully convolutional neural machine translation without this bottleneck
was presented in \cite{extendedngpu,bytenet2016}.
The model in \cite{extendedngpu} (Extended Neural GPU) used
a recurrent stack of gated convolutional layers, while
the model in \cite{bytenet2016} (ByteNet) did away with
recursion and used left-padded convolutions in the decoder.
This idea, introduced in WaveNet \cite{wavenet2016} and also
used in MultiModel (see above) significantly improves efficiency.
Depthwise separable convolutions were first studied by Sifre \cite{sifre2013} and later they were used to get good results
on large-scale image classification with Xception \cite{xception2016}.

\section{Experiments} \label{sec-experiments}

We implemented the MultiModel architecture described above using TensorFlow
and trained it in a number of configurations. In all training runs reported
below we used the same set of hyper-parameters and the Adam optimizer \cite{adam}
with gradient clipping. We will release the implementation as open-source together
with the details of our setup and all used hyper-parameters.
We focused our experiments so as to answer the following questions:
\begin{enumerate}[(1)]
\item How far is the MultiModel trained on 8 tasks simultaneously from state-of-the-art results?
\item How does training on 8 tasks simultaneously compare to training on each task separately?
\item How do the different computational blocks discussed above influence different tasks?
\end{enumerate}

In answering the above questions, we don't always consider all 8 problems.
Especially the 4 translation problems behave very similarly, so we decided to
not include them all in each comparison but we focused on the more varied problems instead.

To answer question (1), we compare the performance of the 8-problem MultiModel with
state-of-the-art results in Table~\ref{tab-sota}. We did not invest much time yet in
tuning hyper-parameters of the MultiModel, so we believe that the difference seen
there will become much smaller with more tuning. The results we achieve are
similar to the ones task-specific models get without heavy tuning, e.g.,
on English-French translation we improve on the Extended Neural GPU
results reported last year \cite{extendedngpu}.

\begin{table}
  \centering
  \begin{tabular}{lcc}
    \toprule
    Problem & MultiModel (joint 8-problem) & State of the art \\
    \midrule
    ImageNet (top-5 accuracy)  & 86\% & 95\% \\
    WMT EN $\to$ DE (BLEU)          & 21.2 & 26.0 \\
    WMT EN $\to$ FR (BLEU)          & 30.5 & 40.5 \\
  \bottomrule
  \end{tabular}\\[4mm]
  \caption{Comparing MultiModel to state-of-the-art from \cite{SzegedyIV16} and \cite{moe}.}
  \label{tab-sota}
\end{table}

To answer question (2), we compare the MultiModel trained jointly
with MultiModel trained separately just on a single task.
When training jointly on 8 tasks, we had a separate worker training on each
task with shared parameters of the model. When training on a single task,
we used only a single worker training on this task for a similar number of steps.
Since we are comparing different instantiations of the same model,
we report two internal metrics: the negative log-perplexity and per-token
accuracy (measured on the development set). As can be seen from the results
in Table~\ref{tab-compare}, the joint 8-problem model performs similarly
to single-model on large tasks, and better, sometimes significantly,
on tasks where less data is available, such as parsing.

\begin{table}
  \centering
  \begin{tabular}{lccccc}
    \toprule
    \multirow{2}{*}{\vspace{-2mm}Problem} & \multicolumn{2}{c}{Joint 8-problem} & &  \multicolumn{2}{c}{Single problem} \\
    \cmidrule{2-3}\cmidrule{5-6}
    & log(perpexity) & accuracy & & log(perplexity) & accuracy \\
    \midrule
    ImageNet   & 1.7 & 66\% & & 1.6 & 67\% \\
    WMT EN$\to$DE & 1.4 & 72\% & & 1.4 & 71\% \\
    WSJ speech & 4.4 & 41\% & & 5.7 & 23\% \\
    Parsing    & 0.15 & 98\% & & 0.2 & 97\% \\
  \bottomrule
  \end{tabular}\\[4mm]
  \caption{Comparison of the MultiModel trained jointly on 8 tasks and separately on each task.}
  \label{tab-compare}
\end{table}

The large improvement on parsing seen in Table~\ref{tab-compare} is not
that surprising taking into account the large number of text data in translation tasks.
But we were curious if training parsing just with ImageNet, a seemingly unrelated task,
would also bring any improvements. This is indeed the case, as can be seen in Table~\ref{tab-parse}.
The difference in performance is significant, and since we use both dropout and early stopping,
we conjecture that it is not related to over-fitting. Rather, it seems, there are computational
primitives shared between different tasks that allow for some transfer learning even between
such seemingly unrelated tasks as ImageNet and parsing.

\begin{table}
  \centering
  \resizebox{\columnwidth}{!}{
  \begin{tabular}{lccccccccccc}
    \toprule
    \multirow{2}{*}{\vspace{-2mm}Problem} & \multicolumn{3}{c}{Alone} & &  \multicolumn{3}{c}{W/ ImageNet} & &  \multicolumn{3}{c}{W/ 8 Problems}\\
    \cmidrule{2-4}\cmidrule{6-8}\cmidrule{10-12}
    & log(ppl) & acc. & full & & log(ppl) & acc. & full & & log(ppl) & acc. & full \\
    \midrule
    Parsing  & 0.20 & 97.1\% & 11.7\% &  & 0.16 & 97.5\% & 12.7\% &  & 0.15 & 97.9\% & 14.5\% \\
  \bottomrule
  \end{tabular}}\\[4mm]
  \caption{Results on training parsing alone, with ImageNet, and with 8 other tasks.
    We report log-perplexity, per-token accuracy, and the percentage of fully correct parse trees.}
  \label{tab-parse}
\end{table}

To answer question (3), we check how training without the mixture-of-experts layers
or without the attention mechanism influences performance on different problems.
Since both these mechanisms were designed with machine translation in mind, we
check the English-French translation. But we also include ImageNet, since this is
the problem that stands the least to benefit from those blocks. In fact, one could
expect that removing these blocks will improve performance on ImageNet alone if they were
truly useless for this task. In contrast, we see in Table~\ref{tab-blocks} that these blocks
either don't affect or slightly improve performance. This leads us to conclude that mixing
different computation blocks is in fact a good way to improve performance on many various tasks.

\begin{table}
  \centering
  \resizebox{\columnwidth}{!}{
  \begin{tabular}{lcccccccc}
    \toprule
    \multirow{2}{*}{\vspace{-2mm}Problem} & \multicolumn{2}{c}{All Blocks} & &  \multicolumn{2}{c}{Without MoE} & &  \multicolumn{2}{c}{Without Attention} \\
    \cmidrule{2-3}\cmidrule{5-6}\cmidrule{8-9}
    & log(perpexity) & accuracy & & log(perplexity) & accuracy & & log(perplexity) & accuracy \\
    \midrule
    ImageNet   & 1.6 & 67\% & & 1.6 & 66\% & & 1.6 & 67\% \\
    WMT EN$\to$FR & 1.2 & 76\% & & 1.3 & 74\% & & 1.4 & 72\% \\
  \bottomrule
  \end{tabular}}\\[4mm]
  \caption{Ablating mixture-of-experts and attention from MultiModel training.}
  \label{tab-blocks}
\end{table}

\section{Conclusions}

We demonstrate, for the first time, that a single deep learning model
can jointly learn a number of large-scale tasks from multiple domains.
The key to success comes from designing a multi-modal architecture in which
as many parameters as possible are shared and from using computational blocks
from different domains together. We believe that this treads a path towards interesting
future work on more general deep learning architectures, especially since our model
shows transfer learning from tasks with a large amount of available data to ones
where the data is limited.

\end{document}